\documentclass[preprint,12pt]{elsarticle}

\usepackage{amssymb}
\usepackage{amsmath}
\usepackage{booktabs}
\usepackage{subcaption}
\usepackage[usenames]{color}

\journal{}

\begin{document}

\begin{frontmatter}



\title{Autoencoder-driven Spiral Representation Learning for Gravitational Wave Surrogate Modelling}


\author[inst1]{Paraskevi Nousi}

\affiliation[inst1]{organization={Department of Informatics, Aristotle University of Thessaloniki},%Department and Organization
            city={Thessaloniki},
            country={Greece}}
            
\author[inst1]{Styliani-Christina Fragkouli}

\author[inst1]{Nikolaos Passalis}

\author[inst3]{Panagiotis Iosif}

\author[inst2]{Theocharis Apostolatos}

\author[inst3]{George Pappas}

\author[inst3]{Nikolaos Stergioulas}

\author[inst1]{Anastasios Tefas}

\affiliation[inst2]{organization={Department of Physics, University of Athens},%Department and Organization
            city={Athens},
            country={Greece}}
            
\affiliation[inst3]{organization={Department of Physics, Aristotle University of Thessaloniki},%Department and Organization
            city={Thessaloniki},
            country={Greece}}

\begin{abstract}
Recently, artificial neural networks have been gaining momentum in the field of gravitational wave astronomy, for example in surrogate modelling of computationally expensive waveform models for binary black hole inspiral and merger. Surrogate modelling yields fast and accurate approximations of gravitational waves and neural networks have been used in the final step of interpolating the coefficients of the surrogate model for arbitrary waveforms outside the training sample. We investigate the existence of underlying structures in the empirical interpolation coefficients using autoencoders. We demonstrate that when the coefficient space is compressed   to only two dimensions, a spiral structure appears, wherein the spiral angle is linearly related to the mass ratio.  Based on this finding, we design a spiral module with learnable parameters, that is used as the first layer in a neural network, which learns to map the input space to the coefficients. The spiral module is evaluated on multiple neural network architectures and consistently achieves better speed-accuracy trade-off than baseline models. A thorough experimental study is conducted and the final result is a surrogate model which can evaluate millions of input parameters in a single forward pass in under 1ms on a desktop GPU, while the mismatch between the corresponding generated waveforms and the ground-truth waveforms is better than the compared baseline methods. We anticipate the existence of analogous underlying structures and corresponding computational gains also in the case of spinning black hole binaries.
\end{abstract}


%\item Using Autoencoders we uncover a latent spiral formation in gravitational waveforms
%\item A novel spiral module is proposed which can be added to any neural network
%\item The addition of the spiral module leads to smaller errors and allows lighter networks to be trained 


\end{frontmatter}


\section{Introduction}
\label{sec:introduction}

% start with gravitational waveform surrogate modelling, importance
% role of neural networks as fitting functions
% approaches so far ignore any hints of underlying structure
% in this work we use autoencoders first to uncover meaningful latent representations
% then model the result as a neural spiral module with 4 parameters which can be used as the first layer of any neural architecture and consistently provides faster convergence, better mismatches as well as more energy-efficient models

The first direct detection of gravitational waves (GW) from a binary black hole (BBH) merger in 2015 \cite{Abbott:2016blz}, initiated the era of Gravitational Wave Astronomy. The Advanced LIGO \cite{TheLIGOScientific:2014jea} and Advanced Virgo \cite{acernese2014advanced} laser interferometric detectors, with arms spanning a few kilometers, can observe stellar-mass and intermediate-mass binary black hole mergers, as well as binary mergers in which one or two components are neutron stars. With consecutive sensitivity improvements, the number of detections has increased from just 3 during the first observing run (O1) \cite{TheLIGOScientific:2016pea}, to 11 at the end of the second observing run (O2) (comprising the first Gravitational-Wave Transient Catalog (GWTC-1) \cite{LIGOScientific:2018mvr}) and to a total 50 events at the end of the first half of the third observing run (O3a) (comprising GWTC-2 \cite{Abbott:2020niy}). 
Furthermore, the international network of GW detectors is expanding with the addition of the KAGRA detector \cite{akutsu2019kagra} (which already participated in O3) and LIGO-India \cite{2021arXiv210501716S} (expected to join in a few years from today). This development is expected to bolster detection rates and  sky-localization accuracies, even more so as the instruments are expected to be upgraded beyond their initial design sensitivity \cite{abbott2020prospects}.

These discoveries were also enabled by tremendous efforts in modelling of GW sources and data analysis (see e.g. \cite{mehta2017, mehta2019, khan2020, estelles2020, cotesta2018, nagar2018, nagar2020, rifat2020, varma2019, khan2019phenom, varma2019precess, williams2020, ossokine2020, dietrich2019tidal, dietrich2019improved, pratten2020domharm, london2018, nagar2020teob, pratten2021IMRPhenomXPHM, garciaquiros2020,LIGOScientific:2020stg, Abbott:2020khf}). 
For non-eccentric BBH mergers, the parameter space $\boldsymbol{\lambda}$ is seven-dimensional, comprising the mass ratio and the three spin components for each black hole. The  numerical-relativity-calibrated effective-one-body (EOB) GW model for binary black hole coalescence
SEOBNRv4 \cite{bohe2017improved} only uses three parameters, assuming spins aligned with the orbital angular momentum. In the EOBNRv2 \cite{pan2011inspiral} model, the black holes are non-spinning and thus only the mass ratio parameterizes the waveform generation routine.

Waveform generation specifically is of particular importance, as balancing the trade-off between computational speed and waveform faithfulness is a challenging task, with a lot of room for improvements. Towards this end, surrogate modelling \cite{field2014fast} can be deployed, wherein the goal is to build a fast surrogate model, which can approximate the waveforms generated from a computationally slower GW model, within a given tolerance. 
The surrogate model proposed in \cite{field2014fast} for the EOBNRv2 model  consists of three steps: First, given a training set of waveforms, a reduced-order-modeling (ROM) basis is obtained (using a greedy algorithm \cite{clrsalgorithms}), such that the training set can be reconstructed by multiplying this basis with corresponding coefficients to within a preset error. Second, given the reduced basis, further compression along the time axis is performed, referred to as Empirical Interpolation Method (EIM)  \cite{barrault2004eim,maday2009eim}. Finally, an interpolation method is used, to obtain fits of the EIM basis coefficients for arbitrary waveforms, which are outside the training set. In practice, such a surrogate model can run significantly faster than required for the generation of the original waveforms.

Further improvements in the computational speed for generating waveforms with a surrogate model were demonstrated in \cite{khan2021gravitational}, who used artificial neural networks (ANNs) in the third step, in place of traditional interpolation methods. This is an important development, since  surrogate-model-parameter fitting with machine-learning (ML) methods has clear computational advantages over traditional methods, especially as the number of physical parameters on which the waveforms depend increases beyond a few (see \cite{setyawati2020regression} for a comparison of various interpolation methods for waveform modeling). More generally, ML has emerged as a robust approach for solving a range of problems in GW astronomy, with a rapid increase in use, see e.g. \cite{cuoco2020review} for a recent review. 

ANNs pose several advantages over traditional interpolation methods for this task. First, the dimensionality of the input space does not pose as much of a challenge, so long as the training set is sufficiently large and the input space is densely sampled \cite{khan2021gravitational}. Second, by leveraging high processing power GPUs, ANNs are able to process hundreds of thousands or even millions of input samples in a single forward pass, depending on the specific architecture. This can lead to extraordinarily large speedups in comparison to traditional interpolation approaches, which are already much faster than evaluating fiducial models \cite{field2014fast}. Finally, as universal approximators \cite{sonoda2017neural}, ANNs are capable of modelling complex relationships, if they have sufficient depth and width.

In this work, we focus on non-spinning  EOBNRv2 waveforms and thus work with a 1-dimensional input space, consisting of only the mass ratios between the two BBH components. First, we utilize Autoencoders \cite{nousi2017deep}, to investigate the existence of any underlying structure in the coefficients to be fitted. We find that by compressing the coefficient space into only 2 dimensions, {\it a spiral structure appears, wherein the spiral angle is linearly related to the mass ratio}. Based on this finding, we design a spiral module with learnable parameters, to be used as the first layer in a neural network, which learns to map the input space to the coefficients. As this structure occurs naturally in a fully unsupervised scenario, we hypothesize, and experimentally show, that the addition of the spiral module leads to faster convergence of the network, as well as to waveform generation that is more faithful to the fiducial waveforms than baseline networks without the module. As a demonstration, we construct a surrogate model for the EOBNRv2 model, which is valid in mass ratios in the interval from 1 to 8 and which can generate up to 1.6 million coefficients in a single forward pass on a desktop GPU, with a worst-case mismatch of $4.33\times 10^{-7}$ (several orders of magnitude lower than the mismatches reported in \cite{bohe2017improved}). The existence of the underlying structure, points towards the possible existence of an analogous (higher-dimensional) underlying structure also in the case of spinning BBH mergers, with corresponding anticipated computational improvements.

The remainder of this paper is structured as follows: Section~\ref{sec:related} discusses previous related works on surrogate modelling for GW waveforms, as well as autoencoders and representation-driven neural networks. In Section~\ref{sec:proposed}, an introduction into surrogate modelling is given first, before moving on to autoencoders and finally to the spiral module. Section~\ref{sec:experimental} is an experimental study into the benefits of the proposed method under various conditions. Finally, Section~\ref{sec:conclusion} concludes this study and summarizes its findings.

\section{Related Work}
\label{sec:related}

A comprehensive framework for surrogate modelling of GW models was given in \cite{field2014fast}. Although the method described is model-agnostic, an effective-one-body (EOB) waveform family was used as an example, namely the EOBNRv2 \cite{pan2011inspiral} model, where the input space is 1-dimensional and specifically only the mass ratio is used to generate waveforms. A training set of waveforms is generated first, in a predetermined input space for which the resulting surrogate is valid. Given a sufficiently large training set, the surrogate model should be able to generalize and approximate waveforms outside of the training set. A validation or test set is generated to showcase the surrogate's ability to generalize.

The proposed framework in \cite{field2014fast} consists of three steps. First, given the training set, a linear decomposition problem is solved, resulting to the generation of a reduced basis and corresponding reduced basis coefficients. A greedy algorithm was proposed for this decomposition which iteratively selects waveforms from the training set to add to the basis. Second, the reduced basis is further reduced with regard to its time dimension. Specifically, time nodes are selected with the Empirical Interpolation Method, which can be used to reconstruct the entirety of the reduced basis. This process results in a modified basis and empirical interpolation coefficients. The final step is to fit the coefficients to the input space, so that the model can be evaluated at arbitrary points in the same interval as the training input space. Since the input space is 1-dimensional, traditional interpolation methods can be used with high accuracy, and it was shown that the resulting surrogate models were significantly faster than the baseline EOBNRv2 generator, while generating waveforms with low mismatches to the ground truth waveforms. Other interpolation methods were compared against each other in \cite{setyawati2020regression}.

In \cite{chua2019reduced}, the waveforms are decomposed into their real and imaginary parts and two surrogate models are created, while the input space for their model is 4-dimensional. ANNs were used to fit the reduced basis coefficients, skipping the empirical interpolation step altogether. Recently, in \cite{khan2021gravitational}, the waveforms were decomposed into their amplitude and phase signals and two surrogate models are created. The GW model used is the SEOBNRv4 model \cite{sonoda2017neural}, whose input space is 3-dimensional. Various neural network architectures are investigated for the fitting task and the corresponding mismatches with the ground truth waveforms are reported. 

We focus our work on the EOBNRv2 model used in \cite{field2014fast}, whose input space is 1-dimensional, and do not decompose the complex waveforms. A complex reduced basis is found, upon which the empirical interpolant is built and thus the final empirical interpolation coefficients are complex. Furthermore, we first investigate the underlying relationship between the input space and the coefficients by utilising Autoencoders \cite{nousi2017deep,nousi2018self} and by modelling the hidden representation which emerges and has a spiral structure with regard to the mass ratio. We then construct neural network architectures instructed by this finding, by designing a spiral module with learnable parameters. To the best of our knowledge, this is the first work to explore the underlying structure of the empirical interpolation coefficients and apply the knowledge gained to a neural network architecture with a novel module, inspired by the uncovered, physical properties of the coefficients themselves. As demonstrated in Section~\ref{sec:experimental}, the added spiral module leads to surrogate waveforms with better mismatches to the ground truth waveforms, while being faster than equivalent architectures, which do not use the module.

\section{Background \& Proposed Method}
\label{sec:proposed}

\subsection{Gravitational wave surrogate models}
\label{sec:surrogate}

Let $h(t;\boldsymbol{\lambda}) = h_+(t;\boldsymbol{\lambda}) - ih_{\times}(t;\boldsymbol{\lambda})$ denote a complex gravitational wave strain \cite{maggiore}, as given by a fiducial model, where $t$ denotes time and $\boldsymbol{\lambda}$ denotes the intrinsic parameters. The intrinsic parameters are in general 7-dimensional (mass ratio and two spins with arbitrary orientation) for inspiraling black holes in general relativity, on non-eccentric orbits, but can be simplified under certain restrictions to only include, for example, the mass ratio and two spins aligned with the orbital plane.

The aim of surrogate modelling is to build an approximation of the signal, denoted as $h_{s}(t;\boldsymbol{\lambda})$, such that $h_{s}(t;\boldsymbol{\lambda}) \approx h(t;\boldsymbol{\lambda})$ within a threshold of error tolerance. If only the dominant, quadrupole ($l=m=2$) mode \cite{maggiore} of gravitational-wave emission is considered, then the target is $h_{s}(t;\boldsymbol{\lambda}) \approx h_{2,2}(t;\boldsymbol{\lambda})$. 
The first step towards the creation of such a surrogate model is to generate a large set of waveforms and subsequently find a Reduced Basis for this set, using for example a greedy algorithm \cite{field2014fast} or Singular Value Decomposition \cite{blackman2017surrogate}. Following \cite{field2014fast}, for non-spinning Effective-One-Body (EOB) waveforms of the EOBNRv2 fiducial model \cite{pan2011inspiral}, the parameter space is one-dimensional and each waveform is parameterized only by the mass ratio $q$.

Thus, a training set of $N$ waveforms is created\footnote{Waveforms were generated using the \texttt{PyCBC} package \cite{pycbc}.} $\{ h_i(t;q_i) \}_{i=1}^{N}$ where $q=\frac{m1}{m2}$ is the mass ratio of the binary system and limited to a predetermined interval, such as $1\leq q \leq 2$ for example, for which the surrogate model will be valid. The greedy algorithm chooses a set of $m < N$ waveforms (and their corresponding $q$ values), which constitute the reduced basis $\{e_i\}_{i=1}^{m}$ after orthonormalization. The basis is built iteratively such that it reconstructs the training set to within a predetermined tolerance as a linear combination of the basis and projection coefficients $\{c_i(q)\}_{i=1}^{m}$:
\begin{equation}
    h(t;q) \approx \sum_{i=1}^{m} c_i(q) e_i(t).
\end{equation}
The reconstruction error is measured using the inner product between the reconstructed and the training set waveforms, as per \cite{field2014fast}, where the inner product is given by the complex scalar product:
\begin{equation}
    \langle h(\cdot; q_1), h(\cdot; q_2) \rangle = \int_{t_{min}}^{t_{max}} h^{*}(t;q_1) h(t;q_2) dt,
\end{equation}
where the star notation denotes complex conjugation. The norm of a waveform is then given by $\|h(\cdot;q) \| = \sqrt{\langle h(\cdot;q),h(\cdot;q) \rangle}$, and waveforms are assumed to be normalized such that $\|h(\cdot;q) \| = 1$ throughout this paper. The overlap integral of a waveform $h(t;q)$ and its corresponding surrogate model prediction $h_s(t;q)$ is given by:

\begin{align}
    \mathcal{O}(h(\cdot;q),h_s(\cdot;q)) &= \text{Re}\langle h(\cdot;q),h_s(\cdot;q) \rangle \\
    &= 1 - \frac{1}{2} \|h(\cdot;q) - h_s(\cdot;q) \|^2,
\end{align}
and finally, the mismatch between these two waveforms is computed as:
\begin{align}
    \label{eq:mismatch}
    \mathcal{M}(h(\cdot;q),h_s(\cdot;q)) &= 1 - \mathcal{O}(h(\cdot;q),h_s(\cdot;q)) \\
    &= 1 - \text{Re}\langle h(\cdot;q),h_s(\cdot;q) \rangle.
\end{align}
The mismatch between real waveforms and surrogate predictions will be used throughout this paper as a measure of the surrogate's performance.

After the greedy algorithm has converged and the reduced basis has been computed, the second step to surrogate modelling is to recast the problem as one of interpolation in time. Specifically, given a reduced basis $\{e_i(t)\}_{i=1}^{m}$, the Empirical Interpolation Method (EIM) aims to find a set of points in time, or \emph{empirical nodes}, $\{ T_i \}_{i=1}^{m}$, such that if the values of the fiducial waveforms are known only at these points, the entirety of the fiducial waveform may be recovered with high accuracy for arbitrary $q$. The result of the EIM method is a new basis that is obtained by imposing on the reduced basis the constraint that the coefficient values are equal to the values of the fiducial waveforms themselves at the empirical time nodes:
\begin{equation}
    a_j(q) = h(T_j;q),
\end{equation}
for all $q$ in the training set. The goal is to fit a model to the coefficients given the corresponding $q$ values present in the training set, such that the surrogate model can approximate any waveform in the range in which it has been trained. Note that using the EOBNRv2 waveform family allows for the generation of a large training set of $N$ waveforms, leading to dense sampling of empirical interpolation coefficients. Thus, the problem is once again recast as an interpolation problem from input space $q$ to the empirical interpolation coefficients $\mathbf{a}(q)=\{a_j\}_{j=1}^{N}$. The set of $N$ waveforms have been aligned at the peak of their amplitudes and cropped them to the length of the shortest waveform \cite{khan2021gravitational,field2014fast}, such that they have a common duration, i.e., time dimension. Finally, the coefficients to be fitted are equal to the values of the training set of waveforms at the empirical time nodes, leading to a dataset $\mathcal{D}$ of pairs of $q$ values and coefficients $\mathbf{a}(q)$:
\begin{equation}
    \mathcal{D} = \{ (q_i, \mathbf{a}(q_i)) \}_{i=1}^{N}.
\end{equation}

% EOBNRv2 (Field et al ref)

% reduced basis (SVD mention)

% EIM + final "groundtruth" (fitting lambda to EIM coeffs)

% goal: input lambda, output coeffs; use autoencoders to uncover any latent representation to link coefficients to q

\subsection{Autoencoders}

Autoencoders (AE) are a class of unsupervised neural networks which are trained to reconstruct their input, by first mapping it onto an intermediate representation, typically of lower dimension~\cite{vincent2008extracting}, naturally offered for representation learning tasks. In general, an AE consists of an encoding part which maps the input to an intermediate representation, and a decoding part which reconstructs the input and typically has an architecture that is symmetrical to the encoder. The encoder and decoder can consist of multiple layers, each of which is accompanied by learnable parameters, such as the weights and biases of fully connected layers. Let $\mathbf{x} \in \mathbb{R}^{D}$ denote a  $D$-dimensional input vector. Then an autoencoder can be formally defined by its two parts as:
\begin{equation}
    \hat{\mathbf{x}} = f(g(\mathbf{x})),
\end{equation}
where $g(\cdot)$, $f(\cdot)$ are the encoding and decoding functions respectively, and $\hat{\mathbf{x}} \in \mathbb{R}^{D}$ is the network's output, which is trained to approximate the input. The encoding and decoding functions can have symmetrical or asymmetrical architectures, and typically consist of multiple layers of, for example, fully connected layers, convolutional layers and even recurrent modules. A useful operation for the decoder in particular is that of transposed convolution, or that of fractionally-strided convolution, used in practice to increase the spatial dimension of feature maps.

\begin{figure}
    \centering
    \includegraphics[width=0.9\linewidth]{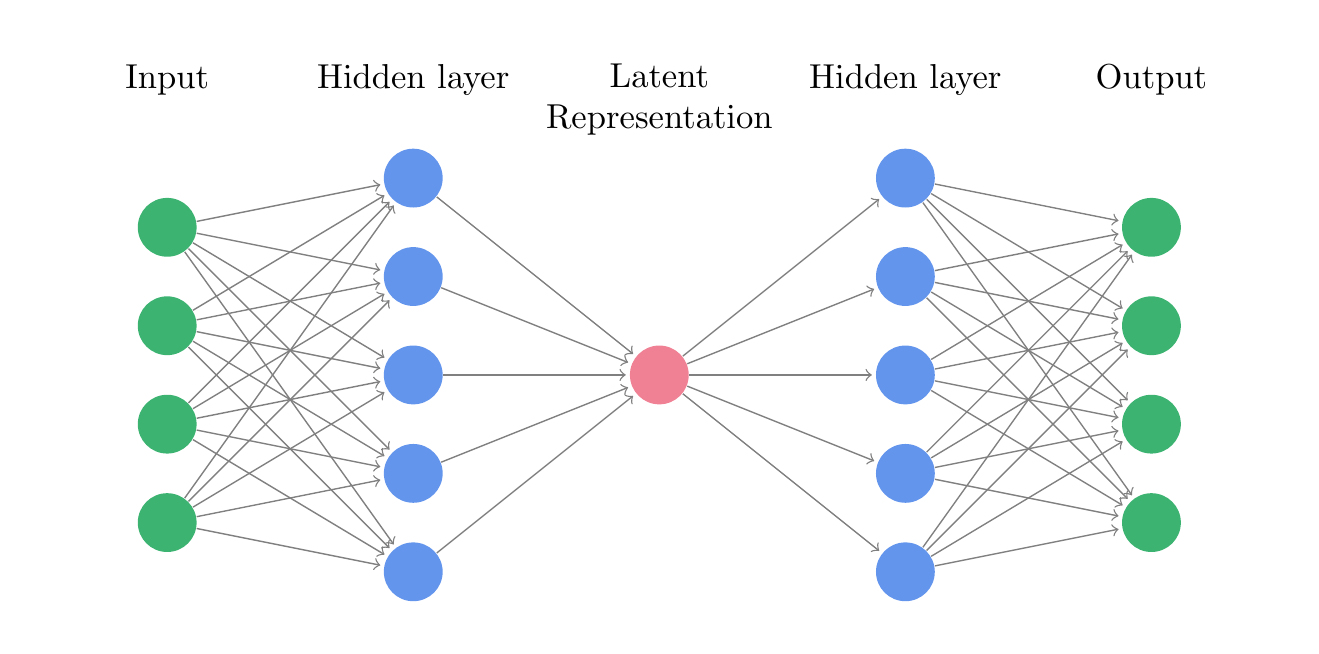}
    \caption{Single hidden layer architecture of a fully connected Autoencoder.}
    \label{fig:ae_architecture}
\end{figure}

Although they are not limited in this scope, typically, AEs are used for dimensionality reduction. In this case, let $\mathbf{y} \in \mathbb{R}^{d}$, $d < D$, denote the output of the encoder, i.e., $\mathbf{y} = g(\mathbf{x})$, then $\mathbf{y}$  may be regarded as a compressed version of $\mathbf{x}$. An AE of this form can be trained by minimizing the Mean Square Error (MSE) between the network's input and output, which corresponds to the reconstruction error:
\begin{equation}
    \mathcal{L} = \frac{1}{N} \sum_{i=1}^{N} \| \hat{\mathbf{x}}_i - \mathbf{x}_i \|_2^2,
\end{equation}
over all training samples, with respect to the network's parameters. As data labels are not taken into consideration, an AE trained using the object described here is fully unsupervised. Figure~\ref{fig:ae_architecture} presents a typical architecture for an AE consisting of $L$ fully connected layers. The input and output layer consist of the same number of neurons $D$. Multiple non-linear layers lead to the intermediate representation. The decoder then tries to reconstruct the input via multiple non-linear layers.

Due to their ability to extract semantically meaningful representations without the use of labels, AEs have been widely studied for a variety of tasks, including clustering~\cite{xie2016unsupervised,nousi2018self}, classification~\cite{nousi2017deep,nousi2017discriminatively}, and image retrieval~\cite{wu2013online,carreira2015hashing}. 

Given an EIM coefficient dataset of mass ratio and coefficients pairs $\mathcal{D}=\{ (q_i, \mathbf{a}_i) \}_{i=1}^{N}$, created as described in Section~\ref{sec:surrogate}, training an autoencoder with the coefficients $\mathbf{a_i}=\mathbf{a}(q_i)$ as its input can implicitly aid in uncovering the hidden relationship between each $q_i$ and the corresponding coefficients. The process is unsupervised as the mass ratios are unknown to the autoencoder. Because the pairing of each $q_i$ to the corresponding coefficients 
$\mathbf{a_i}$ is known beforehand though, it is possible to use the resulting hidden representation to model the relationship between all $q_i$ and $\mathbf{a}_i$. As an example, by setting the intermediate representation dimension to $d=1$, the autoencoder will learn one-dimensional representations $y_i$ for each $\mathbf{a}_i$, associated with the corresponding $q_i$. The goal is then to find a mapping from $q_i$ to $y_i$.

Although this is possible for higher dimensional latent representations as well, in our experiments we set $d=2$ for simplicity, in which case a spiral pattern emerges, when visualizing the hidden representation as a function of $q$. This finding is depicted in Figure~\ref{fig:ae_hidden_rep} and discussed in more detail in Section~\ref{sec:exp_ae}. On this spiral manifold, the mass ratio $q$ and the spiral angle $\theta$ appear to be related in a linear manner, as consecutive angles correspond to consecutive mass ratio values.

\subsection{End-to-end Neural Regression with Learnable Spiral}

Based on the aforementioned observations, we design and propose the use of a neural spiral module, which first transforms the input $q$ into angles $\theta$:
\begin{equation}
    \label{eq:theta}
    \theta = w\cdot q + b,
\end{equation}
and subsequently maps $\theta$ into a spiral structure of the form:
\begin{align}
  \begin{split}
  \label{eq:spiral}
    s_x &= (\alpha + \beta\cdot\theta) \cdot \cos\theta, \\
    s_y &= (\alpha + \beta\cdot\theta) \cdot \sin \theta,
  \end{split}
\end{align}
where $w, b, \alpha$ and $\beta$ are learnable parameters, as the output is differentiable with respect to each of them. Specifically, the partial derivatives are trivially obtained as follows:
\begin{align}
  \begin{split}
  \label{eq:spiral_params_der}
    \frac{\partial s_x}{\partial \alpha} &= \cos\theta, \\
    \frac{\partial s_x}{\partial \beta} &= \theta\cdot \cos\theta,
  \end{split}
\end{align}
(and similarly for $s_y$). Finally, the errors can be back-propagated to the linear transformation layer, as both $s_x$ and $s_y$ are differentiable with respect to $\theta$:
\begin{align}
  \begin{split}
  \label{eq:spiral_theta_der}
    \frac{\partial s_x}{\partial \theta} &= \beta\cdot\cos\theta - (\alpha + \beta\cdot\theta) \sin\theta, \\
    \frac{\partial s_y}{\partial \theta} &= \beta\cdot\sin\theta + (\alpha + \beta\cdot\theta) \cos\theta.
  \end{split}
\end{align}

\begin{figure}
    \centering
    \includegraphics[width=\linewidth]{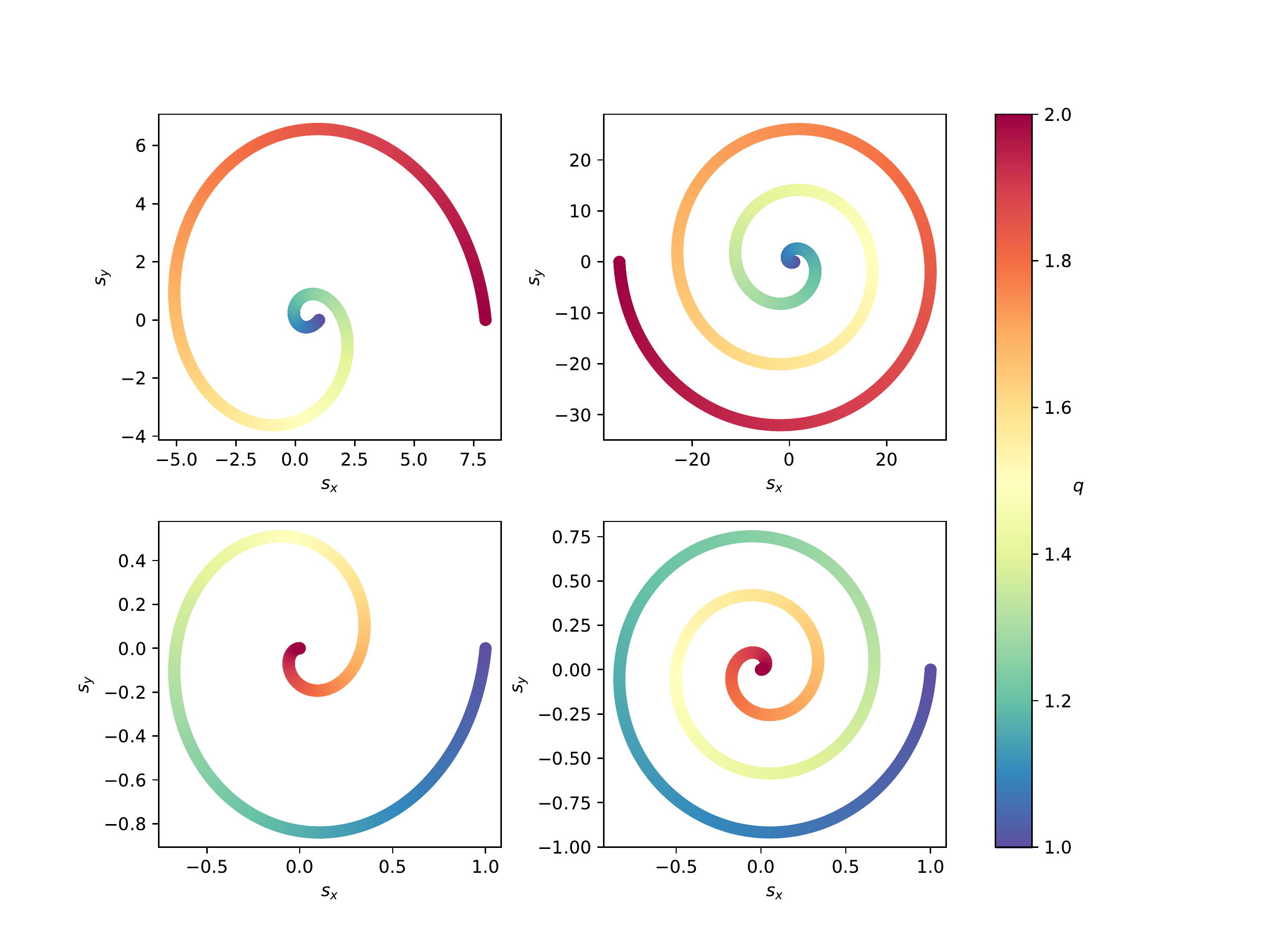}
    \caption{Toy examples of spirals that can be learned with the spiral module in $(s_x, s_y)$ coordinates, given an input $q$ in the range $1\leq q \leq 2$. The spirals are achieved with the following parameter values: (a) $w=-3\pi$, $b=3$, $\alpha=1$, $\beta=3/\pi$, (b) $w=-6\pi$, $b=6$, $\alpha=1$, $\beta=6/\pi$, (c) $w=-3\pi$, $b=3$, $\alpha=1$, $\beta=1/3\pi$, (d) $w=-6\pi$, $b=6$, $\alpha=1$, $\beta=1/6\pi$.}
    \label{fig:spirals}
\end{figure}

\begin{figure}
    \centering
    \includegraphics[width=0.8\linewidth]{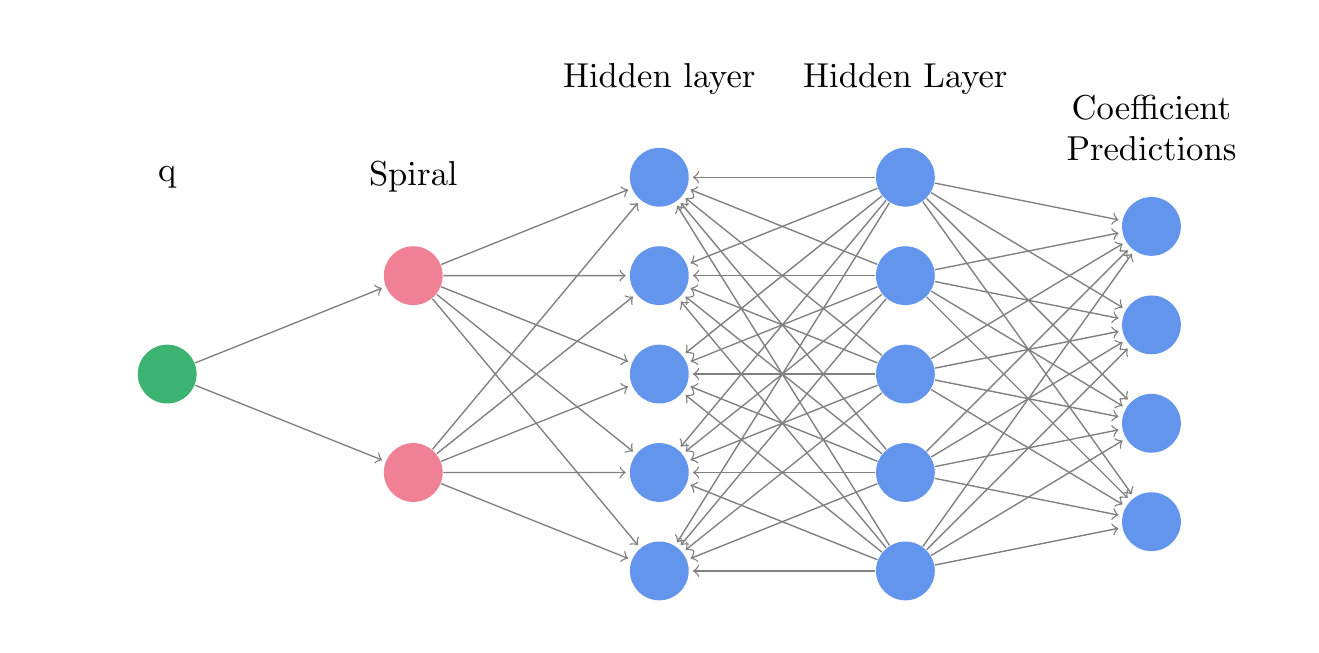}
    \caption{Fully connected neural network with two hidden layers and the proposed spiral module.}
    \label{fig:nn_architecture}
\end{figure}

We hypothesize, and experimentally show in Section~\ref{sec:experimental}, that the addition of this module to a typical neural network helps the convergence of the training process to smaller errors. Figure~\ref{fig:spirals} shows some examples of spirals that can be learned with the spiral module, given an input $q$ in the range $1\leq q \leq 2$. The module can handle various orientations, as well as the degree of coiling. The resulting spiral is then fed to the multiple, subsequent fully-connected layers, each followed by a non-linear activation function, before a final linear layer. An example of this architecture with two hidden layers is shown in Figure~\ref{fig:nn_architecture}.

\section{Experimental Results}
\label{sec:experimental}

\subsection{Autoencoder Representation Learning}
\label{sec:exp_ae}

\begin{figure}
    \centering
    \includegraphics[width=0.8\textwidth]{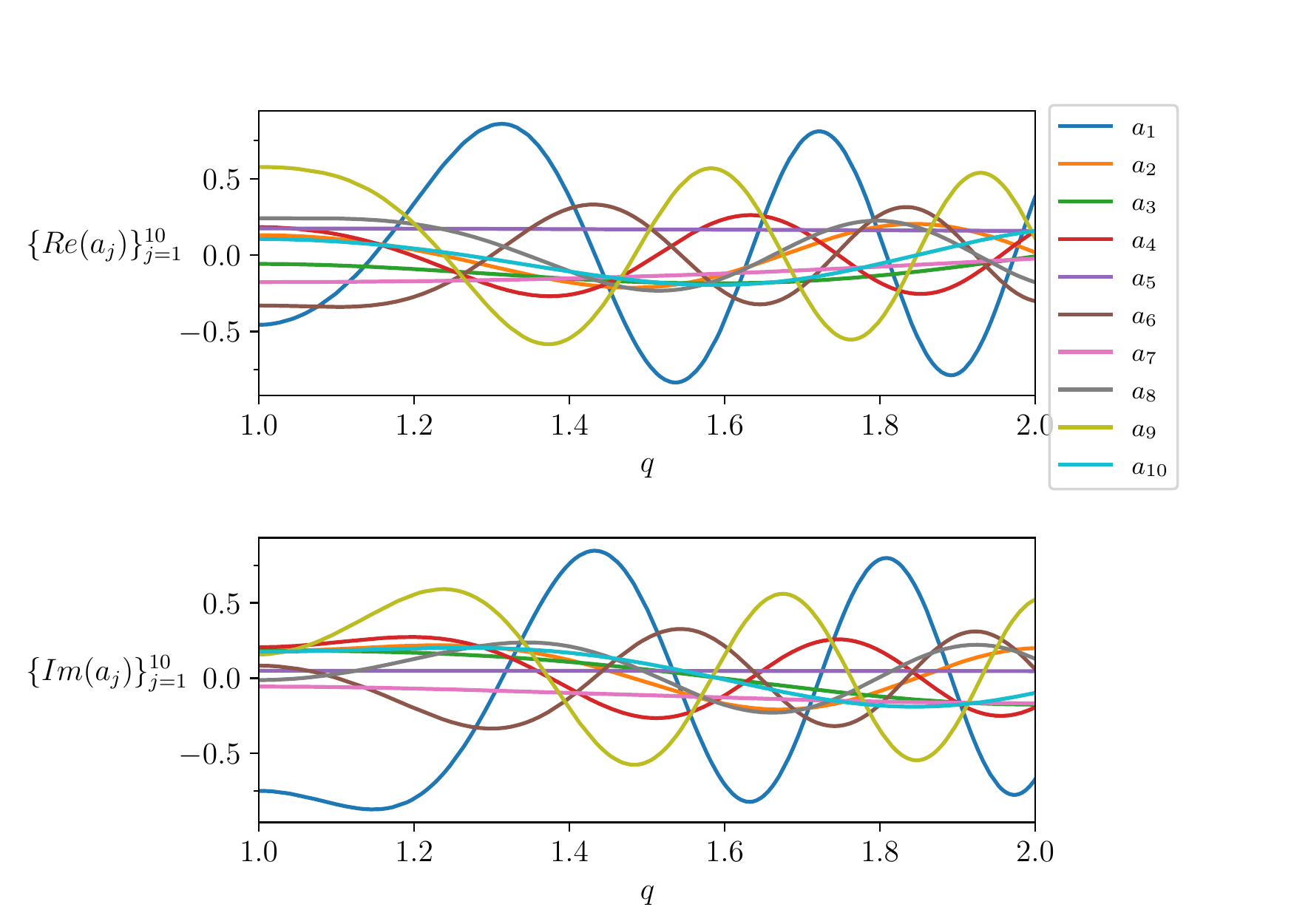}
    \caption{Real (top) and imaginary parts (bottom) of the empirical interpolation coefficients $a_j(q)$ for a surrogate model of EOBNRv2 waveforms that is valid for $1\leq q \leq 2$.}
    \label{fig:coeffs}
\end{figure}

Following \cite{field2014fast}, we first consider non-spinning effective-one-body waveforms with mass ratios in the range $1\leq q \leq 2$. A dataset of $N=1000$ waveforms is generated and a surrogate model is built as described in Section~\ref{sec:surrogate}, with a tolerance of $10^{-10}$, resulting in a reduced basis of size $11$. %\textcolor{red}{Add reference to software packages that were used for this, also in the acknowledgments. .}
The \texttt{RomPy} package \cite{rompy} was used to build the reduced basis and perform the empirical interpolation process. The real and imaginary parts of some of these coefficients along with the corresponding $q$ values are shown in Figure~\ref{fig:coeffs}, where their sinusoidal form relative to $q$ is apparent.

\begin{figure}
    \centering
    \includegraphics[width=0.6\textwidth]{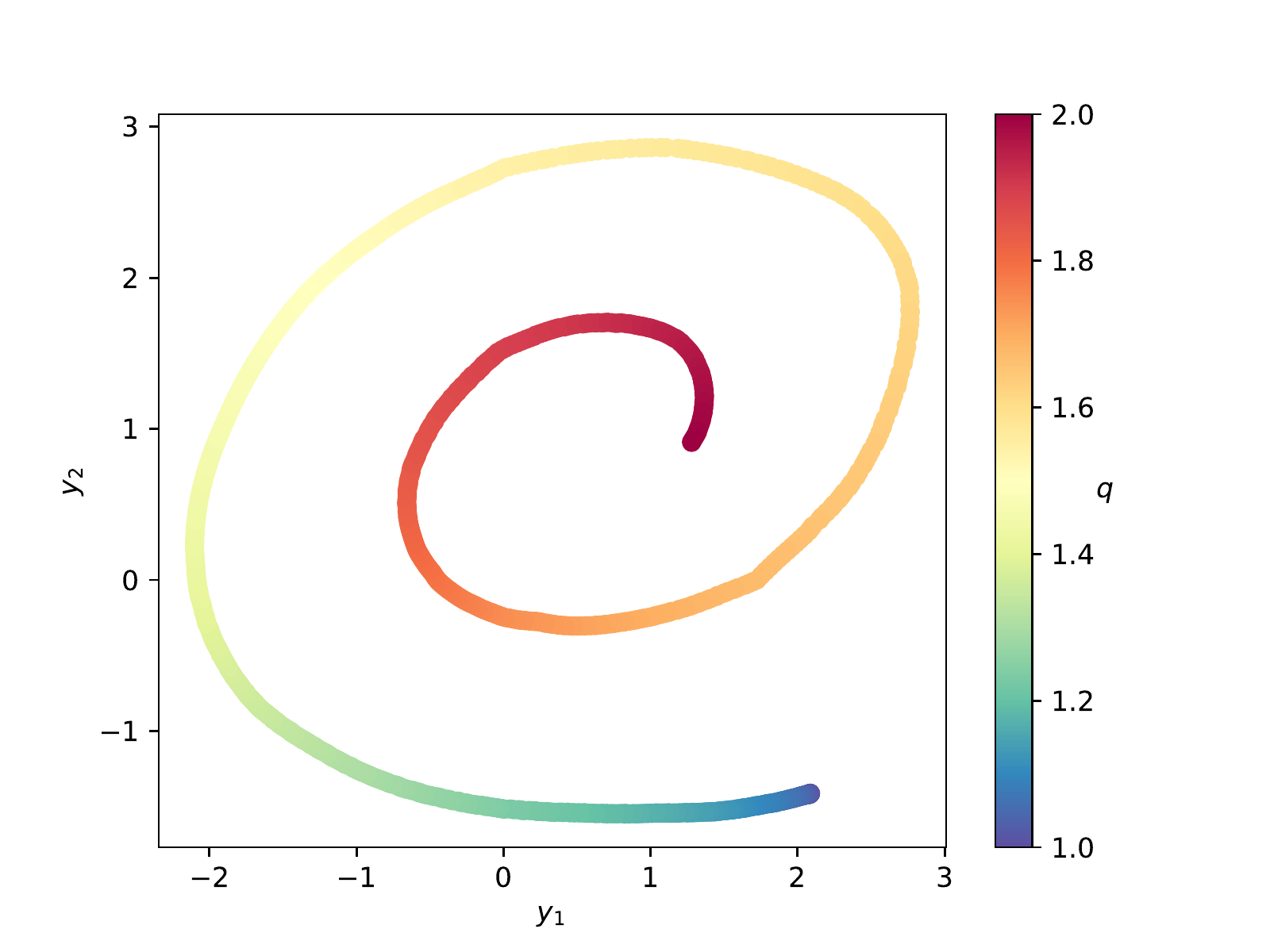}
    \caption{Hidden representation learned by the AE for the empirical interpolation coefficients of a surrogate model of EOBNRv2 waveforms that is valid for $1\leq q \leq 2$.}
    \label{fig:ae_hidden_rep}
\end{figure}

Next, a simple, symmetric encoder-decoder AE architecture is used, with a hidden representation of size $d=2$, and two hidden fully-connected layers of $128$ neurons on either side of the hidden layer. The PReLU non-linearity \cite{he2015delving} was used in every layer. We build our models using the PyTorch Deep Learning framework \cite{pytorch}. The empirical interpolation coefficients were used as both the input and output for this network, with the imaginary parts stacked onto the real parts ($D=2\cdot11=22$). The AE was trained for 100 epochs with a batch size of 32 and an initial learning rate of $0.001$, for which a multiplicative, multi-step schedule was used with a gamma value of 0.9 and a step size of 15. The resulting hidden representation is shown in Figure~\ref{fig:ae_hidden_rep}, where the colormap indicates the corresponding $q$ values for each input coefficient. On the spiral manifold, which presents itself in the hidden layer, the relationship between $q$ and the angle $\theta$ of the spiral appears to be linear. The final reconstruction MSE is $6.82 \times 10^{-5}$.

\begin{figure}
    \centering
    \includegraphics[width=0.6\textwidth]{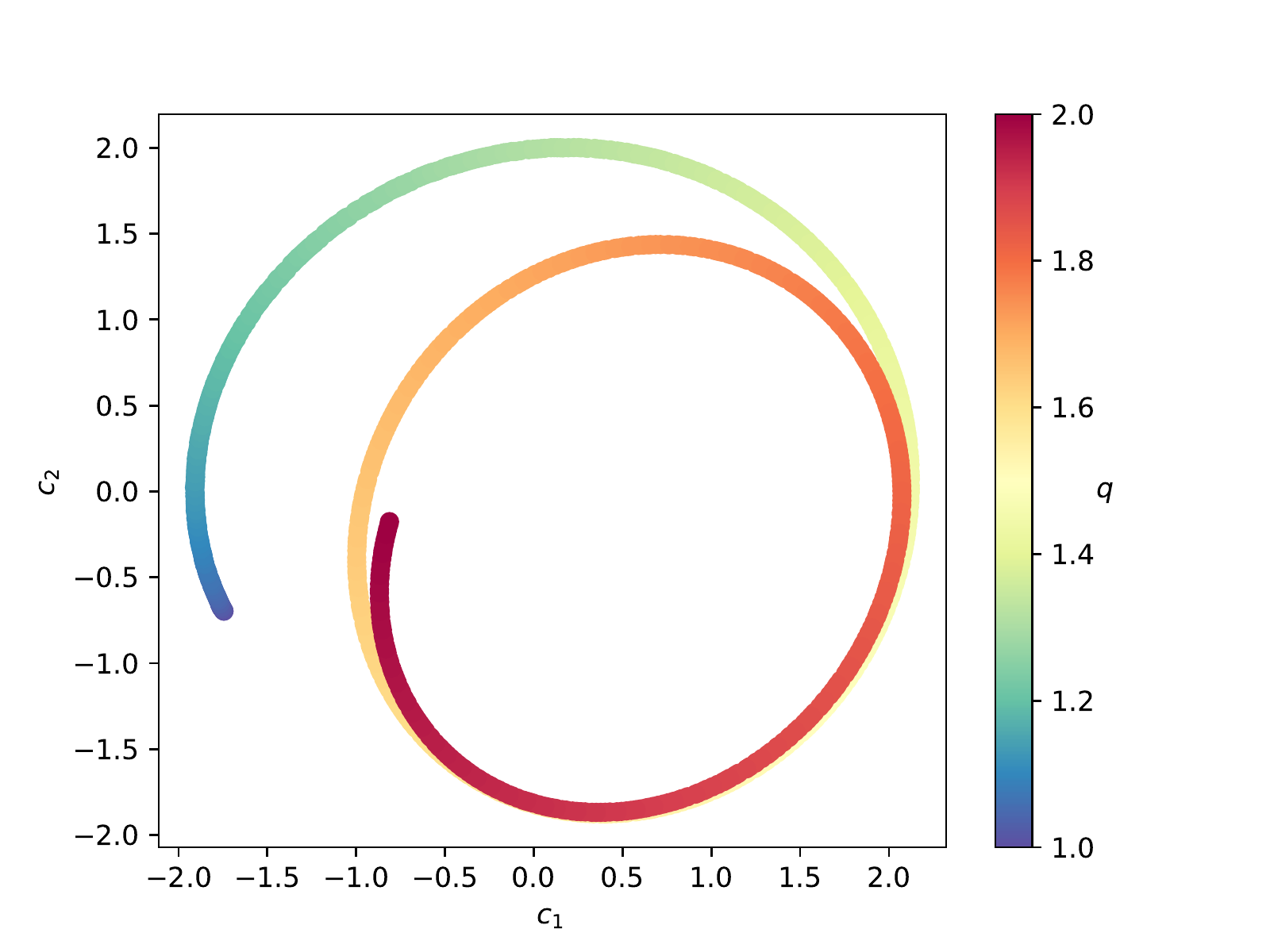}
    \caption{Hidden representation learned by PCA analysis 
    for the empirical interpolation coefficients of a surrogate model of EOBNRv2 waveforms that is valid for $1\leq q \leq 2$.}
    \label{fig:pca_rep}
\end{figure}

To gain insight into the spiral formation that appears, we also perform Principal Component Analysis (PCA) \cite{abdi2010principal} on the same dataset and set the number of principal components to $2$, i.e, $\mathbf{c}_i \in \mathbb{R}^{N}$ for $i=1, 2$. The resulting representation is shown in Figure~\ref{fig:pca_rep}, where a spiral formation also appears. It should be noted that the reconstruction MSE in this case is $3.82 \times 10^{-2}$, that is three orders of magnitude larger than that of the AE described previously. 

\subsection{Learnable Spiral}
% \begin{itemize}
%     \item Best result + corresponding no-spiral net result + time comp
%     \item Architecture search + smallest net w/ spiral for mismatch $\leq 1e-04$
% \end{itemize}

Based on the above observations, we introduce a spiral module, which transforms the input $q$ into angle $\theta$ using Eq.~(\ref{eq:theta}) and subsequently into a spiral using Eq.~(\ref{eq:spiral}). Several neural network architectures with fully-connected layers are evaluated with and without the addition of the spiral module in terms of final waveform mismatch, inference speed as well as their memory requirements, in terms of the maximum batch size that can be processed in a single forward pass on an NVIDIA RTX 2080 Ti GPU. All networks are trained for a total of 2500 epochs, with a batch size of 16 using the Adam optimizer \cite{kingma2014adam} with an initial learning rate of 0.001 which is reduced by 0.95 every 150 epochs.

To obtain a better surrogate model for $1\leq q\leq 2$, a larger training dataset of $N=10000$ waveforms is created with equispaced $q$ values, as well as a validation and a test set, each with $2000$ waveforms with $q$ values sampled uniformly at random in the same range. We first evaluate a traditional spline fitted on the empirical interpolation coefficients, which achieves a minimum, maximum and average mismatch of $1.05 \times 10^{-12}$, $1.26 \times 10^{-8}$, and $1.21 \times 10^{-9}$ respectively.
% TODO: check spline median & 95th percentile
Despite optimizations, the average time needed to produce 10000 waveforms with this method was measured at 0.455s. In contrast, Table~\ref{tab:q1to2_exp} summarizes the worst, median and $95^{\rm th}$ percentile ($p=95$) mismatch achieved by various neural network architectures, as well as the maximum batch size, which can be executed in a single forward pass. Note that the notation $\mathcal{S}$ in the `network' column denotes the insertion of the spiral module, while each number corresponds to the number of neurons per hidden layer. Note also that because of the somewhat lightweight nature of these neural networks, the overhead when predicting millions of coefficients compared to 10000 coefficients is negligible. Specifically, generating the maximum number of coefficients per architecture takes less than 1ms after optimizations.

The addition of the spiral module consistently improves the mismatch achieved. In the case of only one hidden layer, the baseline network with 128 neurons generates waveforms with very poor mismatch ($1.03 \times 10^{-1}$ median mismatch), whereas with the addition of the spiral module even with as few as only 32 hidden neurons, the median mismatch decreases by about 6 orders of magnitude. The best median and $95^{\rm th}$ percentile mismatch ($9.41 \times 10^{-9}$ and $3.48 \times 10^{-8}$) is achieved by the $\mathcal{S}$-32-64-128-64 network, which can generate up to 3.4 million coefficients in a single forward pass on the aforementioned GPU.

\begin{table}[]
    \centering
    \resizebox{0.8\textwidth}{!}{\begin{tabular}{ccccc}
    \toprule
        network & max $\mathcal{M}$ & median $\mathcal{M}$ & $p=95$ $\mathcal{M}$ & max batch size \\ \midrule
        %  128 & 3.76e-01 & 1.03e-01 & 3.20e-01 & 6.1m \\
        %  $\mathcal{S}$-128 & 1.65e-05 & 8.80e-08 & 1.32e-06 & 6.1m \\
        %  $\mathcal{S}$-64 & 1.54e-05 & 2.36e-07 & 1.72e-06 & 9m \\
        %  $\mathcal{S}$-32 & 1.48e-05 & 3.48e-07 & 1.65e-06 & 11.6m \\ \addlinespace
        %   32-64 & 6.86e-05 & 4.92e-07 & 6.13e-06 & 7.3m \\
        %  $\mathcal{S}$-32-64 & 1.69e-06 & 2.93e-08 & 1.35e-07 & 7.3m \\ \addlinespace
        %   32-64-128 & 1.20e-05 & 4.79e-08 & 6.98e-07 & 4.2m \\ 
        %  $\mathcal{S}$-32-64-128 & 1.05e-07 & 3.80e-07 & 7.44e-07 & 4.2m \\ \addlinespace
        %  32-64-128-64 & 1.02e-06 & 4.47e-08 & 1.12e-07 & 3.4m \\ 
        %  $\mathcal{S}$-32-64-128-64 & 1.60e-07 & 9.41e-09 & 3.48e-08 & 3.4m \\
        128 & $3.76 \times 10^{-1}$ & $1.03 \times 10^{-1}$ & $3.20 \times 10^{-1}$ & 6.1m \\
         $\mathcal{S}$-128 & $1.65 \times 10^{-5}$ & $8.80 \times 10^{-8}$ & $1.32 \times 10^{-6}$ & 6.1m \\
         $\mathcal{S}$-64 & $1.54 \times 10^{-5}$ & $2.36 \times 10^{-7}$ & $1.72 \times 10^{-6}$ & 9m \\
         $\mathcal{S}$-32 & $1.48 \times 10^{-5}$ & $3.48 \times 10^{-7}$ & $1.65 \times 10^{-6}$ & 11.6m \\ \addlinespace
          32-64 & $6.86 \times 10^{-5}$ & $4.92 \times 10^{-7}$ & $6.13 \times 10^{-6}$ & 7.3m \\
         $\mathcal{S}$-32-64 & $1.69 \times 10^{-6}$ & $2.93 \times 10^{-8}$ & $1.35 \times 10^{-7}$ & 7.3m \\ \addlinespace
          32-64-128 & $1.20 \times 10^{-5}$ & $4.79 \times 10^{-8}$ & $6.98 \times 10^{-7}$ & 4.2m \\ 
         $\mathcal{S}$-32-64-128 & $1.05 \times 10^{-7}$ & $3.80 \times 10^{-7}$ & $7.44 \times 10^{-7}$ & 4.2m \\ \addlinespace
         32-64-128-64 & $1.02 \times 10^{-6}$ & $4.47 \times 10^{-8}$ & $1.12 \times 10^{-7}$ & 3.4m \\ 
         $\mathcal{S}$-32-64-128-64 & $1.60 \times 10^{-7}$ & $9.41 \times 10^{-9}$ & $3.48 \times 10^{-8}$ & 3.4m \\
         \bottomrule
         
    \end{tabular}}
    \caption{Comparison of various neural network architectures with and without the addition of the spiral module ($\mathcal{S}$) for $1\leq q \leq 2$. $\mathcal{M}$ is the mismatch defined in Eq (\ref{eq:mismatch}). The last column reports the maximum batch size (in millions) that can be processed in a single forward pass on an NVIDIA RTX 2080 Ti GPU .}
    \label{tab:q1to2_exp}
\end{table}

\subsubsection{Extension to larger mass ratios}
% \begin{itemize}
%     \item $q\in (1, 8)$ results w/ spiral + comparison
% \end{itemize}

\begin{figure}
    \centering
    \includegraphics[width=0.8\textwidth]{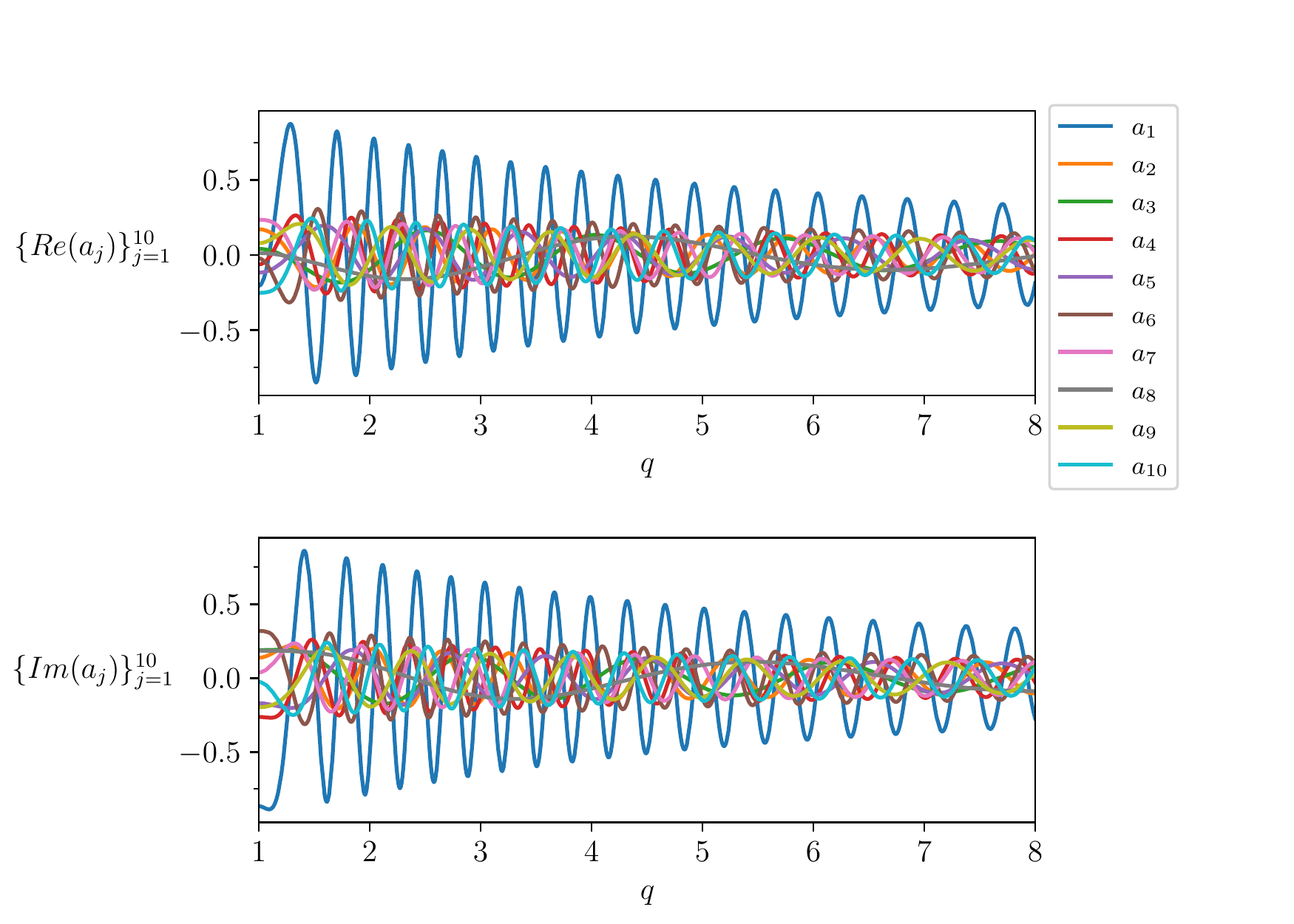}
    \caption{Real (top) and imaginary parts (bottom) of the empirical interpolation coefficients $a_j(q)$ for a surrogate model of EOBNRv2 waveforms that is valid for $1\leq q \leq 8$.}
    \label{fig:coeffs_q1to8}
\end{figure}

We finally build a large training dataset of $N=56000$ waveforms for $1\leq q \leq 8$, with equispaced $q$ values in this range (corresponding to $8000$ values of $q$ per unit interval). A validation and a test set each consisting of $14000$ waveforms are created as well, with $q$ values drawn uniformly at random in the interval $1\leq q \leq 8$ ($2000$ waveforms per unit interval for each set). Figure~\ref{fig:coeffs_q1to8} shows the real and imaginary parts of the first ten coefficients of the EIM basis, $\{a_j(q)\}_{j=1}^{10}$. In spite of some amplitude modulation, each coefficient has a near sinusoidal dependence with $q$ (except near $q=1$, where $dq/da_j=0$ for all $j$).

\begin{figure}
    \centering
    \includegraphics[width=0.9\linewidth]{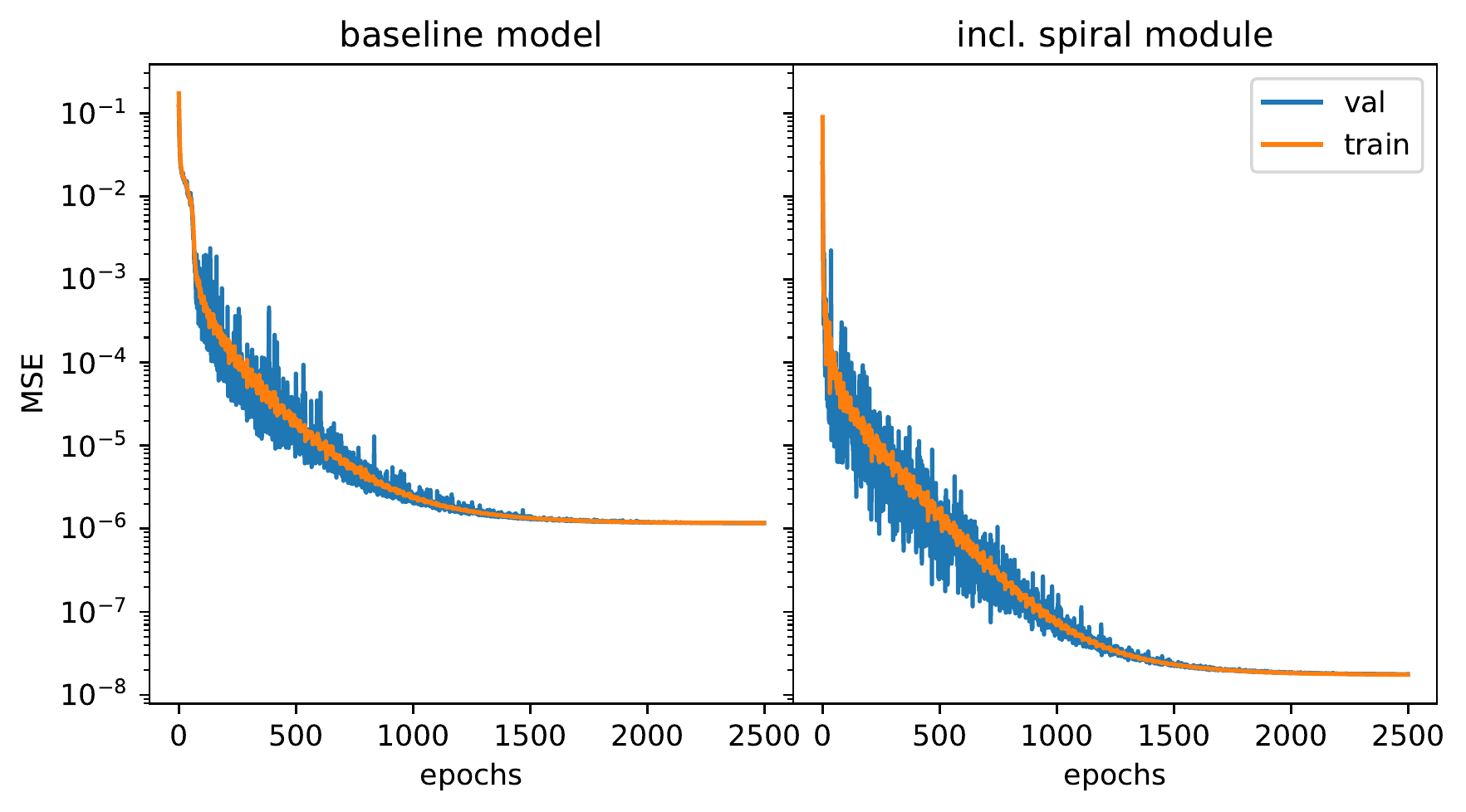}
    \caption{Training and validation loss per epoch for the 32-64-128-64 network and the corresponding architecture with the addition of the spiral.}
    \label{fig:loss}
\end{figure}

Several neural network architectures were trained and evaluated on this dataset. All networks were trained for a total of 5000 epochs, with a batch size of 32 using the Adam optimizer \cite{kingma2014adam} with an initial learning rate of 0.001, which was reduced by 0.9 every 30 epochs. The results are summarized in Table~\ref{tab:q1to8_exp}, in terms of the mismatch and of the maximum batch size that can be used during inference, i.e., the maximum number of coefficients that can be generated in a single forward pass. Note that again, these networks are relatively lightweight and the overhead of generating millions of coefficients versus a hundred thousand coefficients is negligible (specifically, a few microseconds). The training and validation loss per epoch for the $32-64-128-64$ network and the corresponding architecture with the addition of the spiral is shown in Figure~\ref{fig:loss}. The addition of the spiral leads the network to smaller mean squared errors overall. Note also that no overfitting occurs, which can be attributed to the dense sampling of the input space as well as the high sampling rate used during the generation of the training and validation waveforms. Similar loss curves were observed for the rest of the architectures used as well.

\begin{figure}
    \centering
    \includegraphics[width=0.8\textwidth]{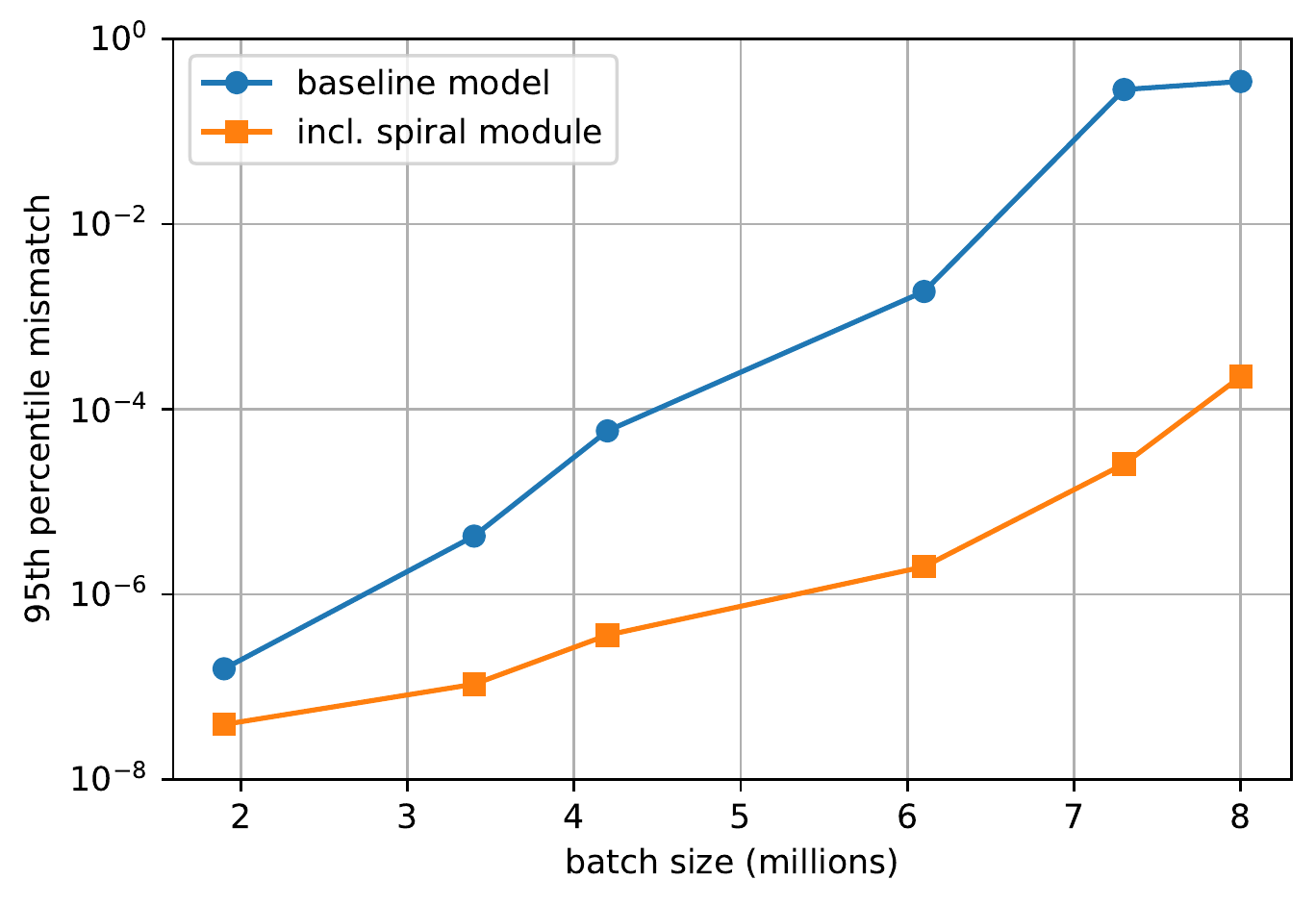}
    \caption{$95^{\rm th}$ percentile mismatch $\cal M$ vs. maximum batch size (in millions) for the surrogate EOBNRv2 model in the $1\leq q \leq 8$ range, where the EIM coefficients were predicted by different baseline neural networks (blue circles) and by corresponding neural networks that included the spiral module (orange boxes), see Table \ref{tab:q1to8_exp}. From left to right, the complexity of the neural network decreases, resulting in a larger mismatch, but also in a larger batch size than can be generated on a particular GPU with a single forward pass.}
    \label{fig:mismatch_vs_batch}
\end{figure}

The maximum batch sizes are also shown in Figure~\ref{fig:mismatch_vs_batch}, as a function of the corresponding $95^{\rm th}$ percentile ($p=95$) mismatches, for the baseline model and for the model that includes the spiral module. As with the $1\leq q \leq 2$ case, the mismatch achieved is consistently better with the addition of the spiral module. Note the case of 7.3m batch size (32-64 network) in particular, where the inclusion of the spiral module achieves a median mismatch of $1.12 \times 10^{-5}$ and a worst case mismatch of $4.39 \times 10^{-3}$, whereas the respective baseline model achieves a median mismatch of $5.85 \times 10^{-3}$ and a worst case mismatch of $4.43 \times 10^{-1}$, about two orders of magnitude worse. Note also that the most lightweight network converges to an undesirable local minimum, whereas the addition of the spiral leads the network to smaller losses and better mismatches.

\begin{table}[]
    \centering
    \resizebox{0.8\textwidth}{!}{\begin{tabular}{ccccc}
    \toprule
        network & max $\mathcal{M}$ & median $\mathcal{M}$ & $p=95$ $\mathcal{M}$ & max batch size \\ \midrule
        16-64 & $4.36 \times 10^{-1}$ & $1.44 \times 10^{-1}$ & $3.47 \times 10^{-1}$ & 8m \\
        $\mathcal{S}$-16-64 & $1.33 \times 10^{-3}$ & $9.14 \times 10^{-5}$ & $2.25 \times 10^{-4}$ & 8m \\ \addlinespace %\hline
        32-64 & $4.43 \times 10^{-1}$ & $5.85 \times 10^{-3}$ & $2.84 \times 10^{-1}$ & 7.3m \\
        $\mathcal{S}$-32-64 & $4.39 \times 10^{-3}$ & $1.12 \times 10^{-5}$ & $2.56 \times 10^{-5}$ & 7.3m \\ \addlinespace
        32-32-64 & $1.48 \times 10^{-3}$ & $4.97 \times 10^{-5}$ & $1.87 \times 10^{-3}$ & 6.1m \\
        $\mathcal{S}$-32-32-64 & $2.99 \times 10^{-5}$ & $9.04 \times 10^{-7}$ & $1.99 \times 10^{-6}$ & 6.1m \\ \addlinespace
        32-64-128 & $9.34 \times 10^{-5}$ & $8.00 \times 10^{-6}$ & $5.83 \times 10^{-5}$ & 4.2m \\
        $\mathcal{S}$-32-64-128 & $1.80 \times 10^{-6}$ & $1.66 \times 10^{-7}$ & $3.63 \times 10^{-7}$ & 4.2m \\ \addlinespace
        32-64-128-64 & $7.79 \times 10^{-5}$ & $1.13 \times 10^{-6}$ & $4.62 \times 10^{-6}$ & 3.4m \\
        $\mathcal{S}$-32-64-128-64 & $2.55 \times 10^{-6}$ & $4.16 \times 10^{-8}$ & $1.07 \times 10^{-7}$ & 3.4m \\ \addlinespace
        64-128-256-128 & $2.46 \times 10^{-5}$ & $1.50 \times 10^{-7}$ & $2.23 \times 10^{-6}$ & 1.9m \\
        $\mathcal{S}$-64-128-256-128 & $5.25 \times 10^{-7}$ & $1.69 \times 10^{-8}$ & $3.93 \times 10^{-8}$ & 1.9m \\ \addlinespace
        \bottomrule
    \end{tabular}}
    \caption{Comparison of various neural network architectures with and without the addition of the spiral module ($\mathcal{S}$) for $1\leq q \leq 8$. The various columns are as in Table \ref{tab:q1to2_exp}.}
    \label{tab:q1to8_exp}
\end{table}

% \subsubsection{Missing data}
% \begin{itemize}
%     \item cases where some $q$ coeffs are missing from training but the spiral helps learn those waveforms better than no spiral
% \end{itemize}

\section{Conclusions}
\label{sec:conclusion}

Recently, artificial neural networks have been gaining momentum in the field of gravitational wave astronomy, and specifically in surrogate modelling of fiducial waveform models. Surrogate modelling yields fast and accurate approximations of gravitational waves and neural networks have been used to interpolate the parameter space to the surrogate coefficients with great success \cite{khan2021gravitational}. Our present work focused on non-spinning Effective-One-Body waveforms of the EOBNRv2 model and we investigated the existence of underlying structures in the empirical interpolation coefficients using autoencoders. Subsequently, the spiral structure that was observed in the latent representation uncovered by the AE, inspired the design of a learnable spiral module. The spiral module can be added to any neural network architecture, \emph{``informing"} the network of the physical structure of the coefficients and resulting to waveforms with better mismatches with respect to the ground truth waveforms. Thus, more lightweight architectures can be used in conjunction with the spiral module to generate millions of coefficients in a single batched forward pass, which can be executed in less than 1ms on a desktop GPU.

The existence of the underlying structure in the case of the 1-parameter family of non-spinning waveforms points towards the possible existence of an analogous (higher-dimensional) underlying structure also in the case of spinning BBH mergers, with corresponding anticipated computational improvements. Our initial investigation of spin-aligned waveforms, indeed confirms this anticipation (in preparation).

\section*{Acknowledgments}

We are grateful to Elena Cuoco, Michael Bejger, Rhys Green, Sebastian Khan and Geraint Pratten for very useful discussions and comments and to the COST network CA17137 “G2Net” for support. The authors gratefully acknowledge the Italian Instituto Nazionale di Fisica Nucleare (INFN), the French Centre National de la Recherche Scientifique (CNRS) and the Netherlands Organization for Scientific Research, for the construction and operation of the Virgo detector and the creation and support  of the EGO consortium. 

 \bibliographystyle{elsarticle-num} 
 \bibliography{main}





% \end{thebibliography}
\end{document}